\documentclass{article}
\PassOptionsToPackage{numbers,sort&compress}{natbib}

\usepackage[preprint]{neurips_2026}

\usepackage[utf8]{inputenc} % allow utf-8 input
\usepackage[T1]{fontenc}    % use 8-bit T1 fonts
\usepackage{url}            % simple URL typesetting
\usepackage{booktabs}       % professional-quality tables
\usepackage{amsfonts}       % blackboard math symbols
\usepackage{amsmath}        % math environments and \text in math mode
\usepackage{nicefrac}       % compact symbols for 1/2, etc.
\usepackage{microtype}      % microtypography
\usepackage{xcolor}         % colors
\usepackage{graphicx}
\usepackage{subcaption}
\usepackage{float}
\usepackage{hyperref}       % hyperlinks (loaded late, as hyperref requires)

\title{Beyond Attention: Signed Integrated Gradients Attribution in a BiomeGPT-Style Microbiome Transformer}

\author{%
  Oren Nelson \\
  Department of Computer Science\\
  University of California, San Diego\\
  San Diego, CA 92122 \\
  \texttt{onelson@ucsd.edu} \\
}

\begin{document}

\maketitle

\begin{abstract}
In a feature-tokenized \cite{ft_transformer_2021} transformer like BiomeGPT \cite{biomegpt2026}, input tokens are constructed by fusing a fixed identity with a sample specific measurement.  In BiomeGPT it is a fixed species and a variable abundance ($T = S + A$). This biological token serves as a unique atomic unit of information within the sample sequence.  In order to understand downstream model classification tasks in feature-tokenized transformers, authors suggest to use the special classification token \texttt{[CLS]}  attention weights \cite{ft_transformer_2021}\cite{devlin2019bert}\cite{biomegpt2026} to rank sample tokens and understand their importance within the sample. However, using \texttt{[CLS]} attention weights suffers from two critical limitations: (1) they are nonnegative, making it impossible to distinguish disease supporting evidence from health supporting evidence \cite{liu2022rethinking}, and (2) they are post token fusion, obscuring the impact on the output of the input sources, i.e., $S$ and $A$.

To address these limitations, we propose to use a signed, fusion-aware attribution method called Integrated Gradients (IG) \cite{sundararajan2017axiomatic}. We further propose a source derived baseline ($T' = S + A_0$) for tabular models \cite{ft_transformer_2021} such as BiomeGPT that preserves species identity as a fixed biological coordinate while isolating the specific effect of abundance variation. Applied to a disease -- health decision margin, this approach yields polarity that explicitly separates pathogenic from protective microbial signals. We demonstrate that this gradient based approach uncovers species-abundance directional relationships on the outcome and sensitivity diagnostics entirely obscured by unsigned \texttt{[CLS]} attention weights.  We further recommend second order Integrated Hessians \cite{janizek2020explaining} in order to understand microbiome community member interaction rules that show how a perturbation in one member affects the derivative with respect to the outcome in another member. For species that produce ambiguous outcomes, this can show which other species drive them toward disease or health along a given abundance level. This work provides a principled approach to explainability in BiomeGPT that is generalizable to other smooth and differentiable feature-tokenized transformer architectures. Code is available at \url{https://github.com/nohren/token-source-attribution}.
\end{abstract}
\section{Introduction}

BiomeGPT is a feature-tokenized transformer foundation model for microbiome data \cite{biomegpt2026}. For each microbial species, a learned species embedding ($S$) is added to an embedding ($A$) of its binned abundance:
\[
T = S + A.
\]
The resulting species--abundance tokens are processed by the transformer, and as in BERT \cite{devlin2019bert} a learned \texttt{[CLS]} token $\in \mathbb{R}^{d}$ provides a sample level representation for downstream tasks such as inflammatory bowel disease (IBD) classification. During classification fine-tuning, the classifier directly consumes the learned \texttt{[CLS]} token, so the classification loss explicitly trains the encoder to route label relevant sample information into that position. In a fully trained classification model, its \texttt{[CLS]} attention weights $\alpha_{cls} \in \mathbb{R}^{s}$ where $s$ is sequence length are often inspected to identify which input tokens are most relevant to the classification decision. 

% During MLM pretraining, \texttt{[CLS]} begins as a shared Gaussian-initialized embedding and has no direct reconstruction target. Nevertheless, it may receive indirect gradient updates if masked-token predictions depend on information passing through \texttt{[CLS]}. 

% In BiomeGPT \cite{biomegpt2026} as well as this work, the \texttt{[CLS]} attention output is averaged across all layers and heads.

 However, relying solely on these raw attention magnitudes to explain model predictions presents a fundamental limitation: the inability to capture impact polarity \cite{liu2022rethinking}. Attention weights are nonnegative and normalized:
\[
\alpha_i \in [0,1],
\qquad
\sum_i \alpha_i = 1.
\]
Because raw attention weights are strictly positive, they only indicate the magnitude of the model's focus but cannot differentiate between a species token that positively contributes to a prediction and one that actively suppresses it. Consequently, a microbiome feature receiving a massive attention weight might actually be pushing the model away from its final IBD classification. Assuming that high attention inherently equates to a positive contribution can lead to a "faithfulness violation," where the derived explanation contradicts the model's true logic \cite{liu2022rethinking}.   This disconnect ultimately motivates our use of signed integrated gradients to resolve the issue by providing directional polarity of feature impacts within BiomeGPT classification decisions.

% Interpreting a microbiome foundation model may reveal more than the features used by a single task specific classifier. Unlike random forests or gradient-boosted trees trained independently for each task, a foundation model learns reusable representations during pretraining and adapts them during disease specific fine-tuning. Attribution may therefore help identify both the microbial evidence used for a prediction and the pretrained structure reused across downstream tasks.

This work studies Integrated Gradients (IG) as a signed, output directed alternative to \texttt{[CLS]} attention weights in BiomeGPT. IG integrates input gradients along a path from a baseline to the observed input \cite{sundararajan2017axiomatic}. Its attributions are signed and satisfy completeness: up to numerical integration error, their sum equals the corresponding change in model output. We attribute the IBD--healthy logit margin:
\[
F(S,A) = z_{\mathrm{IBD}}(S,A) - z_{\mathrm{Healthy}}(S,A).
\]
Positive attribution indicates that an observed abundance increases the IBD margin, whereas negative attribution indicates support for healthy classification. IG therefore connects each species--abundance observation to a directional change in the decision being explained, rather than merely reporting its prominence within an internal attention pattern.

Applying IG to fused tokens requires a meaningful baseline. An all-zero token would remove both the abundance measurement and the species identity through which the model interprets that measurement. We instead treat species identity as a fixed biological coordinate and abundance as the sample-dependent signal. Let
\[
A_0 = \mathrm{MLP}_{AB}(0)
\]
denote the learned zero-abundance embedding. We define the fused-token baseline as
\[
T' = S + A_0,
\]
which gives
\[
T - T' = (S+A) - (S+A_0) = A - A_0.
\]
The resulting attribution explains how moving from zero abundance to the observed abundance changes the IBD--healthy margin while preserving species identity throughout the integration path. For additive fusion, this is equivalent to applying IG directly to the fused token with the source-derived baseline ($T'=S+A_0$). More generally, the construction suggests that feature-tokenized models should preserve the feature-identity component while interpolating the component that encodes the observed value.

The main contributions of this work are as follows:

\begin{enumerate}
    \item We show that \texttt{[CLS]} attention weights provide unsigned, post-fusion token rankings and therefore cannot distinguish evidence supporting IBD from evidence supporting healthy classification.
    \item We formulate explainability as signed attribution of the IBD--healthy decision margin, connecting each observed abundance to its directional effect on the model output.
    \item We derive a fusion-aware IG baseline, $T'=S+A_0$, that attributes abundance variation while preserving species identity as biological context.
    \item We empirically compare IG with \texttt{[CLS]} attention weights in a BiomeGPT-style IBD classifier and analyze path-averaged gradients as a complementary sensitivity diagnostic.
\end{enumerate}

\section{Related Work}
\textbf{Microbiome foundation models and attention-based interpretation.}
Foundation models have recently been adapted across biological domains, including single-cell transcriptomics with models such as scBERT \cite{yang2022scbert}, Geneformer \cite{theodoris2023geneformer}, and scGPT \cite{cui2024scgpt}. These models illustrate how large scale pretraining can learn reusable biological representations for downstream prediction tasks. Microbiome foundation models extend this idea from molecules or cells to community scale microbial ecosystems. Prior microbiome-specific models have explored language-model-style pretraining over microbial communities \cite{pope2025microbiome_language_model}, while BiomeGPT \cite{biomegpt2026} introduced a species level gut microbiome foundation model pretrained on shotgun sequenced MetaPhlAn \cite{blanco2023metaphlan4} species-abundance profiles.

BiomeGPT's input representation combines microbial species identity with binned abundance, making it a natural example of a feature-tokenized biological transformer. Its interpretation analysis uses \texttt{[CLS]} attention scores to identify microbial taxa associated with health and disease. This work builds on that interpretation goal, but argues that attention rankings are insufficient for signed model output interpretation: attention is nonnegative and post-fusion, so it cannot indicate whether a species-abundance token supports IBD or healthy classification.

\textbf{Attention as explanation.}
Transformer explainability often begins with attention weights, especially attention from a final \texttt{[CLS]} token. Prior work has debated whether attention weights should be interpreted as explanations \cite{jain2019attention, wiegreffe2019attention}. This paper does not attempt to resolve that broader debate. Instead, it makes a narrower claim for microbiome transformers: \texttt{[CLS]} attention can rank fused species-abundance tokens, but it cannot provide signed evidence for a disease--health decision margin. Attention can indicate which tokens receive high attention mass, but the attention weight itself does not encode whether a token increases or decreases the target logit contrast.

\textbf{Integrated Gradients and signed attribution.}
Integrated Gradients (IG) was introduced by Sundararajan et al. \cite{sundararajan2017axiomatic} as an axiomatic attribution method satisfying sensitivity, implementation invariance, and completeness. IG attributes a scalar model output by integrating gradients along a path from a baseline input to the observed input. This makes IG attractive for interpreting the IBD--Healthy logit contrast, because positive and negative attributions can be tied to changes in the model's decision margin. Completeness is also useful in this setting: the sum of IG attributions equals the change in model output from baseline to input, up to numerical integration error. The IG computations in this study are implemented using Captum \cite{captum}.

\textbf{Feature-tokenized transformers and fused-token baselines.}
Feature-tokenized transformers map structured inputs into token embeddings before self-attention. FT-Transformer \cite{ft_transformer_2021}, for example, uses a feature tokenizer to map numerical and categorical columns into embeddings, producing one token per feature. These tokens may combine fixed feature identity with sample-specific value information. In a BiomeGPT-style encoder, each token similarly corresponds to a species-abundance pair, with a fixed species embedding and a dynamic abundance embedding composed before the transformer.

This token structure creates a baseline-design problem for IG. Sundararajan et al. argue that a useful baseline should represent an ``absence of signal'' \cite{sundararajan2017axiomatic}, but for a composed token it is ambiguous what should be absent. In additive species-abundance tokenization, each token is \(T=S+A\), where \(S\) identifies the species coordinate and \(A\) encodes the sample-specific abundance. A naive all-zero fused baseline would remove both the abundance signal and the species identity needed to interpret that signal. Therefore, this work derives the fused-token baseline from the token sources:
\[
T' = S + A_0,
\]
where \(A_0\) is the learned zero-abundance embedding.

\textbf{Attribution paths in tokenized models.}
Recent work on Uniform Discretized Integrated Gradients (UDIG) \cite{roy2024udig} highlights that straight-line interpolation in embedding space may be poorly matched to tokenized models, where intermediate points may not correspond to valid inputs. Although this work performs IG in abundance-embedding space, a future extension is to integrate through the learned abundance encoder rather than linearly between endpoint embeddings. This would make the integration path better aligned with the feature tokenizer that maps abundance values into transformer input embeddings.

\textbf{Integrated Hessians and feature interactions.}
Integrated Hessians extend Integrated Gradients from first-order feature attribution to pairwise feature interaction attribution \cite{janizek2020explaining}. Rather than asking only how much a feature contributes to the output, Integrated Hessians ask whether one feature changes the model's sensitivity to another. This is relevant to microbiome transformers because disease evidence may depend not only on individual species abundances, but also on model-learned abundance-abundance interactions. This work focuses on first-order signed attribution; second-order interaction attribution is left for future work.

\section{Problem Setup}

\subsection{Feature tokenizer.}
Given a species ID \(s_i\) and a discrete binned abundance value \(a_i\), the fused token for species \(i\) is
\begin{equation}
    \mathbf{t}_i
    =
    \mathrm{Embed}_{ID}(s_i)
    +
    \mathrm{MLP}_{AB}(a_i).
\end{equation}
The dataset comes from the curated metagenomic data repository \cite{pasolli2017curatedmetagenomicdata} and consists of discrete binned $a_i$ values for every species $s_i$. The input sequence is constructed by prepending a learned \(\texttt{[CLS]}\) token:
\[
T = (\mathbf{t}_{\texttt{[CLS]}}, \mathbf{t}_1, \dots, \mathbf{t}_{1662}).
\]
Each sequence \(T\in\mathbb{R}^{1663\times512}\) represents one microbiome sample. The zero-abundance embedding represents the absence of a species in the sample. No positional encodings are added, since the microbiome is treated as invariant to species order.

\subsection{Model forward pass.}
The transformer encoder maps the fused species-abundance token sequence to contextualized representations:
\[
H = \Phi(T).
\]
The final \(\texttt{[CLS]}\) representation is used for classification:
\[
h_{\texttt{[CLS]}} = H_0,
\qquad
z = g(h_{\texttt{[CLS]}}),
\]
where \(z\in\mathbb{R}^2\) contains the logits for the IBD and Healthy classes. The model is trained with cross-entropy loss on these logits.

\subsection{Integrated Gradients attribution.}
Let \(S\) denote the species embedding sequence and \(A\) denote the abundance embedding sequence, so that the fused input is
\[
T = S + A.
\]
Let \(S'\) and \(A'\) denote the corresponding baseline embeddings. Integrated Gradients defines a path from baseline to input:
\[
S_\alpha = S' + \alpha(S-S'),
\qquad
A_\alpha = A' + \alpha(A-A'),
\qquad
\alpha\in[0,1],
\]
with fused path
\[
T_\alpha = S_\alpha + A_\alpha.
\]

To obtain signed clinical evidence, the scalar attribution target is the IBD--Healthy logit contrast:
\[
F(S_\alpha,A_\alpha)
=
z_{\mathrm{IBD}}(S_\alpha,A_\alpha)
-
z_{\mathrm{Healthy}}(S_\alpha,A_\alpha).
\]
Positive attribution values therefore increase the IBD--Healthy margin, while negative values decrease it and support the healthy class.

The forward function receives \(S_\alpha\) and \(A_\alpha\) as separate differentiable inputs, then fuses them before the transformer encoder. The source-wise IG terms are
\[
IG_S
=
(S-S') \odot
\int_0^1
\frac{\partial F(S_\alpha,A_\alpha)}{\partial S}
\,d\alpha,
\qquad
IG_A
=
(A-A') \odot
\int_0^1
\frac{\partial F(S_\alpha,A_\alpha)}{\partial A}
\,d\alpha.
\]
Here, \(\odot\) denotes elementwise multiplication by the baseline-to-input displacement. This formulation matches the multi-input Integrated Gradients implementation used in Captum \cite{captum}.

Integrated Gradients satisfies completeness: the sum of the attributions equals the change in model output from baseline to input, up to numerical integration error:
\[
\sum_k IG_k(x)
=
F(x)-F(x').
\]
In this setting, completeness links the attribution scores directly to the model's IBD--Healthy logit difference.

\subsection{Token-fusion baseline for signed Integrated Gradients attribution}
\label{sec:baseline_choice}
The previous sections motivate Integrated Gradients (IG) as a signed alternative to \texttt{[CLS]} attention. Applying IG to BiomeGPT-style species--abundance tokens, however, requires a meaningful baseline. Sundararajan et al. recommend choosing a baseline that represents an ``absence of signal'' \cite{sundararajan2017axiomatic}. For a fused BiomeGPT token, the central question is therefore: which component(s) constitutes the signal to be removed?

BiomeGPT constructs each token through additive fusion:
\[
T = S + A,
\]
where
\[
S = \mathrm{Embed}_{ID}(s_i),
\qquad
A = \mathrm{MLP}_{AB}(a_i).
\]
The species embedding ($S$) identifies the biological feature, whereas the abundance embedding ($A$) encodes its sample-specific value. This decomposition is analogous to feature tokenization in tabular Transformers. Gorishniy et al., for example, construct numerical feature tokens as \cite{ft_transformer_2021}
\[
T_j = b_j + f_j(x_j),
\]
where $b_j$ is a learned feature-specific term and $f_j(x_j)$ encodes the observed value. The first term identifies the tabular coordinate being measured, while the second specifies the observation at that coordinate.

In the BiomeGPT-style architecture, $S=\mathrm{Embed}_{ID}(s_i)$ similarly defines a fixed biological coordinate, while $A=\mathrm{MLP}_{AB}(a_i)$ represents the observed abundance. Because samples share a fixed species vocabulary, the species identities remain constant across samples while their abundance values vary. The abundance embedding is therefore the dynamic, sample specific signal, and the species embedding provides the fixed context in which that signal is interpreted.

\begin{figure}[H]
\centering
\includegraphics[width=1\textwidth]{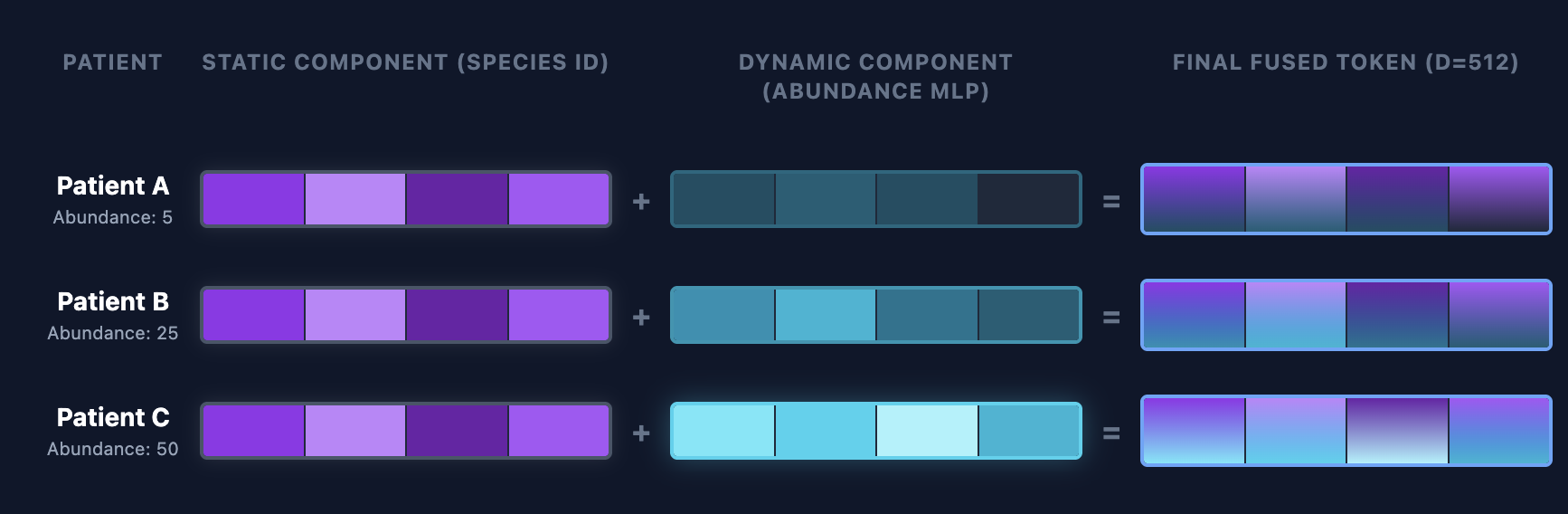}
\caption{The species embedding remains fixed, while the abundance embedding varies across samples.}
\label{fig:my_image}
\end{figure}

An all-zero fused-token baseline,
\[
T'=\mathbf{0}\in\mathbb{R}^{512},
\]
is mathematically valid and is evaluated in Section~\ref{sec:experiments}. However, it removes both the observed abundance and the identity of the biological feature being measured forcing the model to potentially evaluate out of distribution. We instead derive the baseline from the model's tokenization function by preserving species identity and replacing only the abundance component with its learned zero-abundance embedding:
\[
A_0 = \mathrm{MLP}_{AB}(0),
\]
\[
S'=S,
\qquad
A'=A_0.
\]
The observed and baseline fused tokens are therefore
\[
T=S+A,
\qquad
T'=S'+A'=S+A_0.
\]
Their displacement is
\[
T-T' = (S+A)-(S+A_0) = A-A_0.
\]
Species identity cancels from the displacement but remains present in the model computation. The fused-token interpolation path is
\[
T_\alpha = T'+\alpha(T-T') = S+A_0+\alpha(A-A_0).
\]
Equivalently, the source-wise paths are
\[
S_\alpha = S'+\alpha(S-S') = S,
\qquad
A_\alpha = A_0+\alpha(A-A_0).
\]
Thus, the species embedding remains fixed throughout the forward path, while the abundance embedding moves from its learned zero-abundance state to its observed value.

\begin{figure}[H]
\centering
\includegraphics[width=1\textwidth]{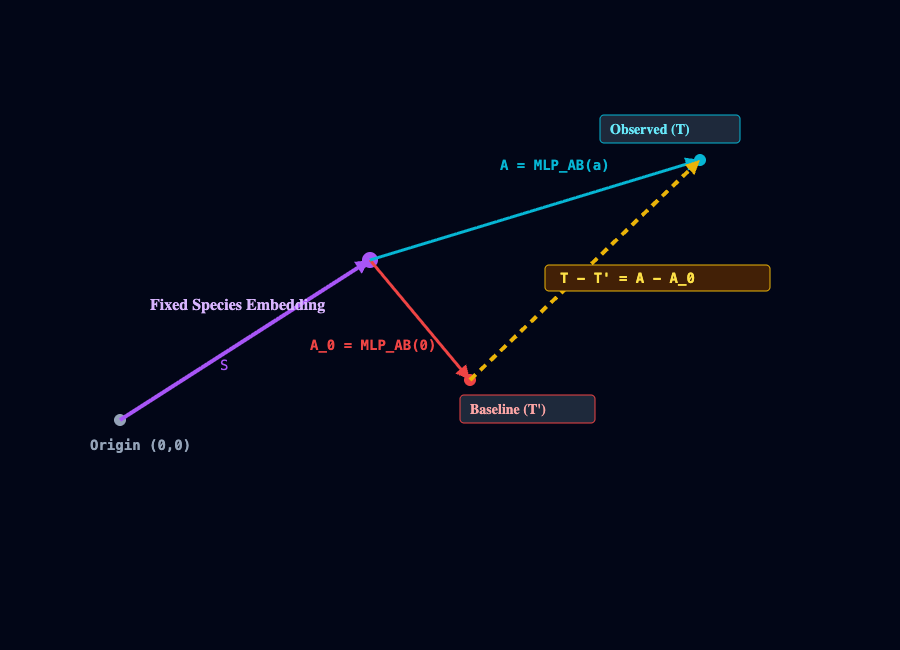}
\caption{The yellow dotted line represents the integration steps from $A_0$ to $A$. Species identity remains throughout the path, providing fixed biological context during gradient attribution.}
\label{fig:manifold_cancellation}
\end{figure}

The corresponding source-specific IG terms are
\[
IG_S = (S-S')\odot \int_0^1 \frac{\partial F(S_\alpha,A_\alpha)}{\partial S} \,d\alpha,
\]
\[
IG_A = (A-A')\odot \int_0^1 \frac{\partial F(S_\alpha,A_\alpha)}{\partial A} \,d\alpha.
\]
Because $S'=S$, species attribution is zero by construction:
\[
IG_S=0.
\]
This does not imply that the model ignores species identity. The species embedding remains in every forward pass and conditions the gradients at each point along the path. Rather, the attribution assigns displacement-weighted evidence to the abundance component while treating species identity as fixed context.

One might nevertheless ask whether setting $IG_S=0$ discards a distinct path-integrated gradient signal. Under additive fusion,
\[
T=S+A,
\]
both sources enter the fused representation through the identity map:
\[
\frac{\partial T}{\partial S}=I,
\qquad
\frac{\partial T}{\partial A}=I.
\]
The chain rule therefore gives
\[
\frac{\partial F}{\partial S} = \frac{\partial F}{\partial T},
\qquad
\frac{\partial F}{\partial A} = \frac{\partial F}{\partial T}.
\]
When evaluated along the same interpolation path, the path-averaged source gradients are consequently identical:
\[
G_S = \int_0^1 \frac{\partial F(S_\alpha,A_\alpha)}{\partial S} \,d\alpha,
\qquad
G_A = \int_0^1 \frac{\partial F(S_\alpha,A_\alpha)}{\partial A} \,d\alpha,
\]
with
\[
G_S=G_A.
\]
The source-specific IG terms differ only through their displacement factors:
\[
IG_S=(S-S')\odot G_S,
\qquad
IG_A=(A-A')\odot G_A.
\]
Thus, the zero species attribution follows from holding species identity fixed, not from the absence of species-conditioned gradient information.

IG may equivalently be applied directly to the fused token, provided that its baseline is derived from the underlying sources. Using $T'=S+A_0$, fused-token IG becomes
\[
\begin{aligned}
IG_T
&= (T-T') \odot \int_0^1 \frac{\partial F(T_\alpha)}{\partial T} \,d\alpha \\
&= (A-A_0) \odot \int_0^1 \frac{\partial F(S,A_\alpha)}{\partial A} \,d\alpha \\
&= IG_A.
\end{aligned}
\]
Therefore, under additive fusion, fused-token IG and abundance-source IG are equivalent when the fused-token baseline is $T'=S+A_0$. The source-level decomposition remains important because it establishes the semantic meaning of the baseline: the resulting attribution explains the effect of moving abundance from its zero state to its observed value while preserving species identity as context.

This construction is not specific to BiomeGPT. It generalizes to feature-tokenized tabular transformers \cite{ft_transformer_2021} in which a fixed feature identity embedding is additively fused with a sample dependent value embedding. The chosen baseline preserves the fixed identity feature coordinate component and interpolate only the component encoding the observed value. To the best of our knowledge, prior applications of IG to such architectures have not been explicitly derived or documented.

More generally, additive fusion causes the source wise gradients to share the same adjoint, leaving displacement as the only distinction between $IG_S$ and $IG_A$. In architectures based on concatenation, gating, FiLM conditioning, or other nonlinear fusion mechanisms, both the source-wise gradients and their appropriate baselines may differ. Source-specific IG must therefore be defined and interpreted relative to the model's token-fusion equation and architecture.

\section{Experiments}
\label{sec:experiments}

To evaluate signed attribution as an alternative to \texttt{[CLS]} attention, this work reimplements a BiomeGPT-style microbiome transformer encoder and applies fusion-aware Integrated Gradients to its species-abundance tokens. The experimental goal is not to establish state-of-the-art IBD classification performance. Instead, the goal is to test whether Integrated Gradients can recover signed disease-versus-health evidence from a trained microbiome transformer, and whether this evidence provides information unavailable from standard \texttt{[CLS]} attention rankings.

The model was pretrained with a masked abundance reconstruction objective on approximately 27,000 MetaPhlAn-profiled \cite{blanco2023metaphlan4} stool samples from the curated metagenomic data repository \cite{pasolli2017curatedmetagenomicdata}. A downstream classification head was then fine-tuned on a balanced IBD-versus-healthy cohort of 8,000 samples, with 4,000 IBD samples and 4,000 healthy samples. The classifier achieved a validation accuracy of approximately 0.93, indicating that the model learned an IBD-discriminative signal suitable for attribution analysis.

Two choices of baseline were compared: the source derived baseline \(T'=S+A_0\) and an experimental all-zero "absence of signal" baseline $T'=0 \in \mathbb{R}^{512}$. For each batch, I monitored the Integrated Gradients convergence delta returned by Captum, which measures the numerical gap between the sum of attributions and the model-output difference \(F(x)-F(x')\). In preliminary runs, the source derived baseline produced substantially smaller Captum convergence deltas than an all-zero fused baseline under the same integration settings. This suggests that the source derived path is both semantically better aligned with the tokenization and also numerically easier to integrate in this model. A systematic comparison of convergence error across baselines is left for future work.

All training and attribution experiments were run on NVIDIA H100 GPUs. Because this study is a proof of concept, I do not evaluate generalization on an external held-out dataset. Instead, the experiments focus on explainability: whether the proposed IG baseline produces signed abundance evidence from the model's latent representation, and how that evidence differs from unsigned \texttt{[CLS]} attention.

\section{Results}
Across 7,382 evaluated samples, abundance source attribution values had mean \(-0.0011\) and ranged from \(-32.77\) to \(36.83\). True Positive samples had a positive mean attribution value of \(0.0027\), while True Negative samples had a negative mean attribution value of \(-0.0045\). These aggregate signs are consistent with the chosen logit contrast target: positive attribution corresponds to model evidence increasing the IBD margin, while negative attribution corresponds to model evidence decreasing the margin.  

The attribution analysis shows a clear contrast between \texttt{[CLS]} attention and signed Integrated Gradients. Standard attention assigns each unmasked token a nonnegative weight \(\alpha_i \in [0,1]\), with \(\sum_i \alpha_i = 1\). Thus, attention can rank which fused species-abundance tokens the model attends to, but the attention weight itself does not indicate whether a token supports IBD or healthy classification. In contrast, Integrated Gradients is applied to the scalar logit contrast
\[
F(S,A) = z_{\mathrm{IBD}}(S,A) - z_{\mathrm{Healthy}}(S,A),
\]
so positive attribution increases the IBD--Healthy margin, while negative attribution decreases it.

This distinction is especially visible for zero-abundance tokens. \texttt{[CLS]} attention weights can remain nonzero for unmasked tokens even when abundance is zero, because softmax attention assigns positive mass to every unmasked token. IG with the source-derived baseline assigns zero attribution when the observed abundance source equals the zero-abundance baseline \(A_0\), because the displacement term \(A-A_0\) is zero. Thus, attention can indicate participation in the transformer computation, but it cannot by itself identify signed abundance evidence.

\begin{figure}[H]
     \centering
     \begin{subfigure}[b]{\textwidth}
         \centering
         \includegraphics[width=0.99\textwidth]{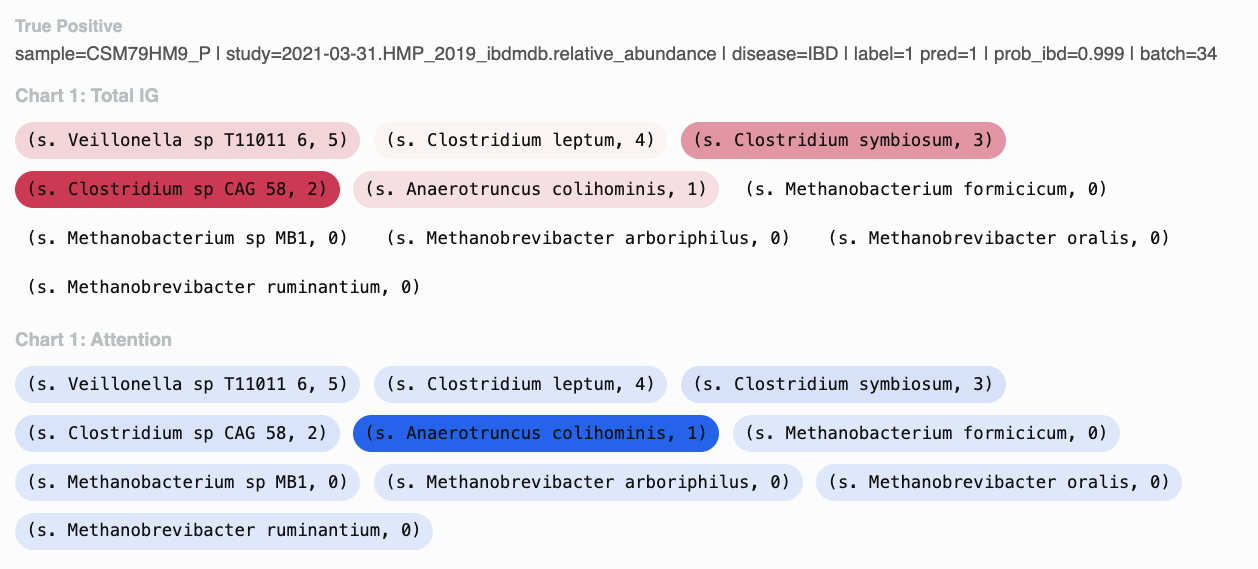}
     \end{subfigure}
     
     \vspace{0.5cm}
     
     \begin{subfigure}[b]{\textwidth}
         \centering
         \includegraphics[width=0.99\textwidth]{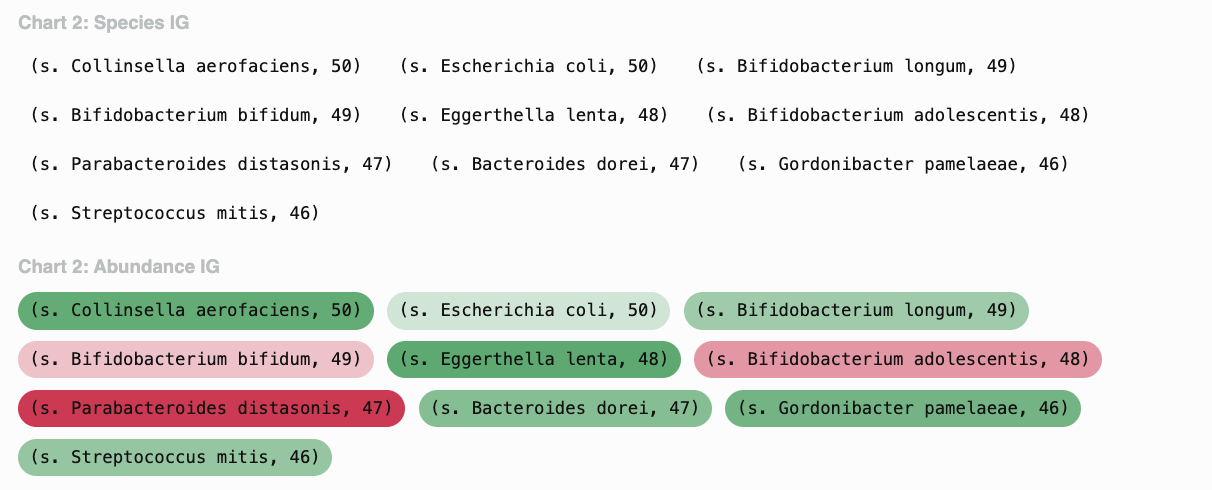}
     \end{subfigure}
     
     \caption{
     Comparison of \texttt{[CLS]} attention weights and Integrated Gradients for species--abundance tokens on a single inference: one true positive sample (\texttt{CSM79HM9\_P}, IBDMDB cohort) classified as IBD with $p_{\mathrm{IBD}}=0.999$. Tokens are annotated (species, binned abundance). In the IG panels, red denotes positive attribution, increasing the IBD--Healthy logit margin; green denotes negative attribution, supporting the healthy class; uncoloured tokens carry zero attribution. Attention (blue) is unsigned. Shade indicates magnitude in all panels. In Chart 1 (low-abundance tokens), attention peaks on \textit{Anaerotruncus colihominis} (abundance 1), whereas the largest attribution driving this prediction belongs to \textit{Clostridium} sp. CAG 58 (abundance 2); the most attended token is not the largest contributor to the decision. Zero-abundance tokens retain nonzero attention mass, as softmax assigns positive weight to every unmasked token, but receive exactly zero IG attribution, since the displacement \(A-A_0\) vanishes. In Chart 2 (high-abundance tokens), species attribution is zero by construction under the baseline \(S'=S\); all signed evidence therefore resides in the abundance source, which separates these species into IBD-supporting and health-supporting contributions.
     }
     \label{fig:attention_vs_ig_zero_abundance}
\end{figure}

\begin{figure}[H]
     \centering

     \begin{subfigure}[b]{0.282\textwidth}
         \centering
         \includegraphics[width=\textwidth]{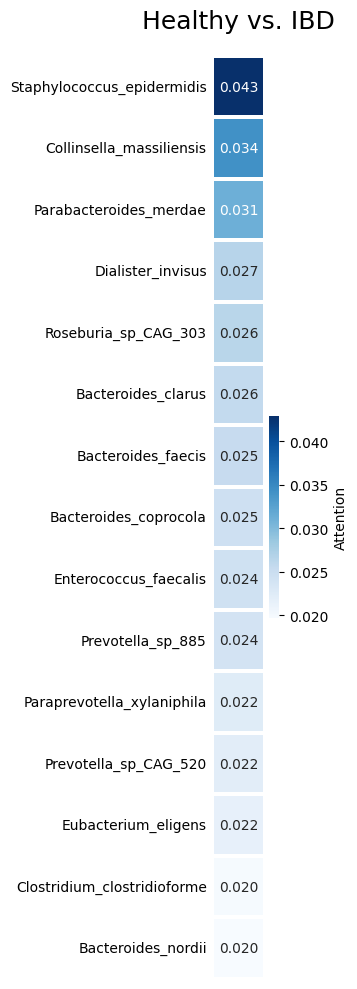}
         \label{fig:attention_top10}
     \end{subfigure}
     \hfill
     \begin{subfigure}[b]{0.33\textwidth}
         \centering
         \includegraphics[width=\textwidth]{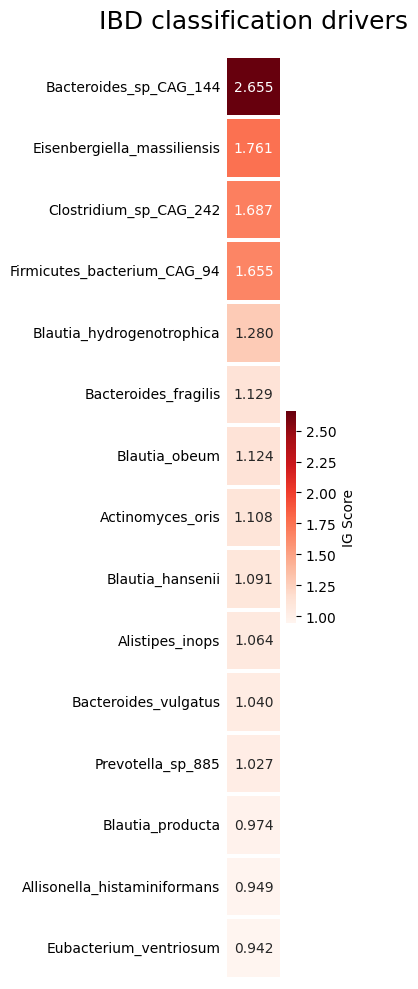}
         \label{fig:ibd_driver}
     \end{subfigure}
     \hfill
     \begin{subfigure}[b]{0.339\textwidth}
         \centering
         \includegraphics[width=\textwidth]{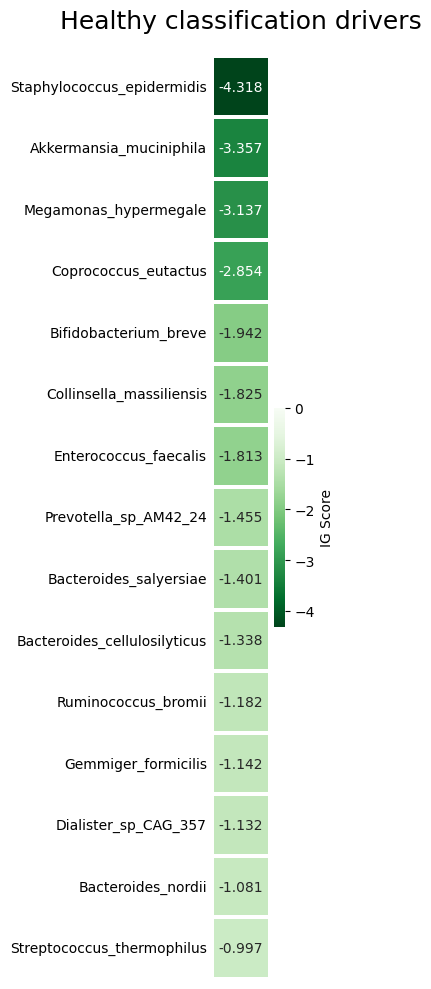}
         \label{fig:health_driver}
     \end{subfigure}

     \caption{
     Top species rankings for IBD classification. Left: species receiving the highest \texttt{[CLS]} attention. Center: species with the largest positive mean IG attribution, increasing the IBD--Healthy logit margin. Right: species with the largest negative mean IG attribution, decreasing that margin and supporting the healthy class. The two species receiving the most attention mass, \textit{Staphylococcus epidermidis} and \textit{Collinsella massiliensis}, both carry negative IG attribution, placing them among the strongest health-supporting species rather than among the IBD drivers. This pattern is not confined to those two: four of the fifteen highest-attention species also rank among the fifteen strongest health drivers, while only one ranks among the strongest IBD drivers. The classifier emits a logit for each class and IG attributes their difference, so the sign of an IG score states which class a species' abundance supports. \texttt{[CLS]} attention is nonnegative and normalized, so a large weight establishes only that the model uses a token, not which of the two classes that use favors.
     }
     \label{fig:clinical_polarity_comparison}
\end{figure}

Figure~\ref{fig:clinical_polarity_comparison} compares top species rankings after excluding species in the bottom 5\% of observed abundance. The remaining species scores are aggregated across all true positive and true negative samples showing \texttt{[CLS]} attention weights and IG. This shows species that consistently receive the highest attention mass, as well as species that consistently have the largest positive or negative IG aggregated.
% Attention gives a single unsigned ranking. IG separates species-abundance tokens into positive evidence increasing the IBD--Healthy margin and negative evidence decreasing it. This is the central explainability improvement over \texttt{[CLS]} attention weights:  the method does not merely rank attended microbes, but assigns directionality to their contribution to the disease-versus-health decision. 

\subsection{Path-Averaged Abundance Sensitivity}

The main comparison in this work is between \texttt{[CLS]} attention and full Integrated Gradients attribution. Attention ranks fused tokens by unsigned attention mass, while IG assigns signed, displacement-weighted evidence to abundance movement along a baseline to input path. Once this IG path is defined, we can also inspect the non-displacement-weighted term inside IG as a complementary sensitivity diagnostic.

For the abundance source,
\[
\mathrm{IG}_{A}
=
(A-A_0) \odot G_A,
\qquad
G_A =
\int_0^1
\frac{\partial F(S,A_\alpha)}{\partial A}
\, d\alpha,
\]
where
\[
A_\alpha = A_0 + \alpha(A-A_0).
\]
The full IG score \(\mathrm{IG}_A\) measures displacement-weighted contribution and satisfies completeness up to numerical integration error. The term \(G_A\), by contrast, omits the displacement multiplier and therefore should not be interpreted as a complete attribution score. This distinction follows the local/global distinction in gradient-based attribution methods discussed by Ancona et al. \cite{ancona2018towards}: Integrated Gradients averages gradients along a baseline to input path, while completeness depends on the full attribution summing to the output difference \(F(S,A) - F(S',A_0)\).

I report \(G_A\) as a complementary abundance-sensitivity diagnostic. Unlike endpoint saliency, \(G_A\) is still averaged along the same baseline-to-input path used by IG. However, because it is not multiplied by \(A-A_0\), it asks a different question: which abundance sources is the model most sensitive to per unit movement in abundance-embedding space? This can highlight species whose abundance branch has high signed model sensitivity even when their observed abundance displacement is small. In this sense, \(G_A\) provides another source-level signal unavailable from \texttt{[CLS]} attention: attention can rank fused tokens, but it does not measure signed sensitivity to abundance movement.

\begin{figure}[H]
     \centering
     \begin{subfigure}[b]{0.4\textwidth}
         \centering
         \includegraphics[width=\textwidth]{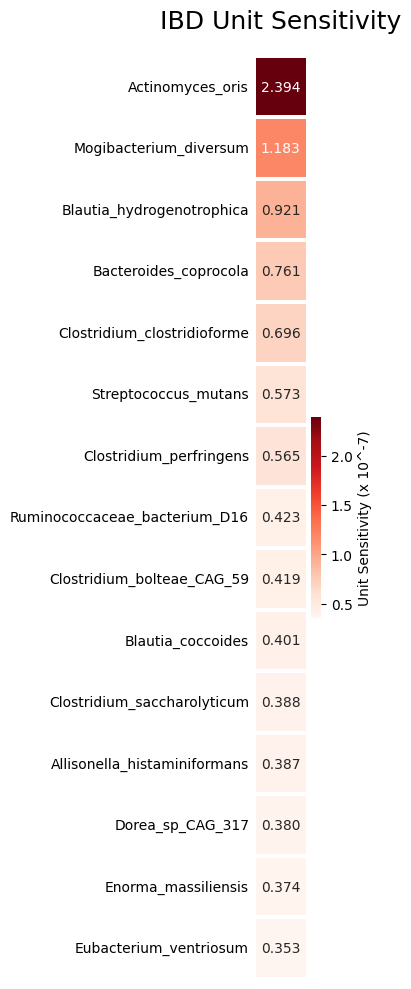}
         \label{fig:ibd_unit}
     \end{subfigure}
     \hfill
     \begin{subfigure}[b]{0.4\textwidth}
         \centering
         \includegraphics[width=\textwidth]{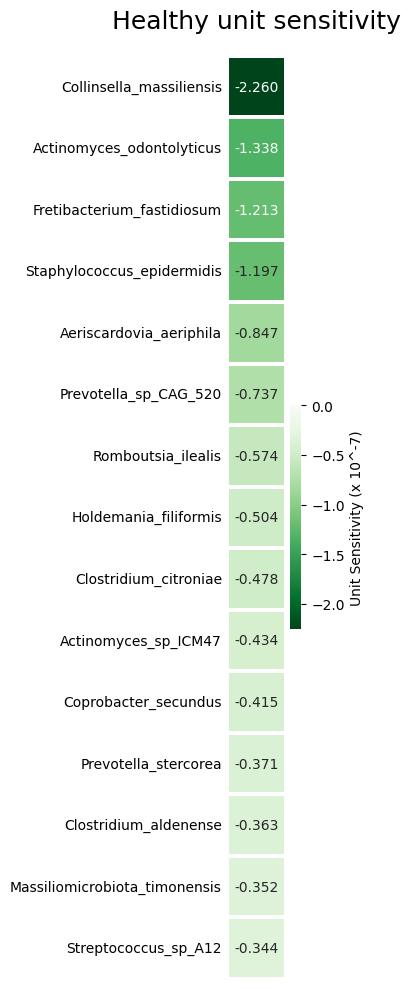}
         \label{fig:health_unit}
     \end{subfigure} 

     \caption{
     Top species ranked by path-averaged abundance sensitivity \(G_A\). Full IG attribution is the product \((A-A_0)\odot G_A\), so a species can rank highly simply because its observed abundance lies far from baseline: the displacement factor scales attribution with abundance, and abundant species are favoured accordingly. \(G_A\) removes that factor and retains only the gradient averaged along the same baseline-to-input path, measuring the change in the IBD--Healthy logit margin per unit of abundance movement. Left: species with the largest positive sensitivity, increasing the margin per unit moved. Right: species with the largest negative sensitivity, decreasing it. Because this ranking is independent of observed abundance, it identifies species to whose abundance the model is highly sensitive but which are present in small quantity, and which full IG therefore ranks lower or omits entirely (Figure~\ref{fig:clinical_polarity_comparison}). These are diagnostics rather than attributions: lacking the displacement factor, they do not satisfy completeness.
     }
     \label{fig:unit_sensitivity}
\end{figure}

\subsection{Abundance--Attribution Response Profiles}

Beyond producing signed species rankings, IG allows attribution to be examined as a function of observed abundance. This provides a second advantage over \texttt{[CLS]} attention: attention ranks fused tokens, but it does not directly show how increasing or decreasing abundance changes the model's disease--health evidence for a given species.

\begin{figure}[H]
\centering
\includegraphics[width=1\textwidth]{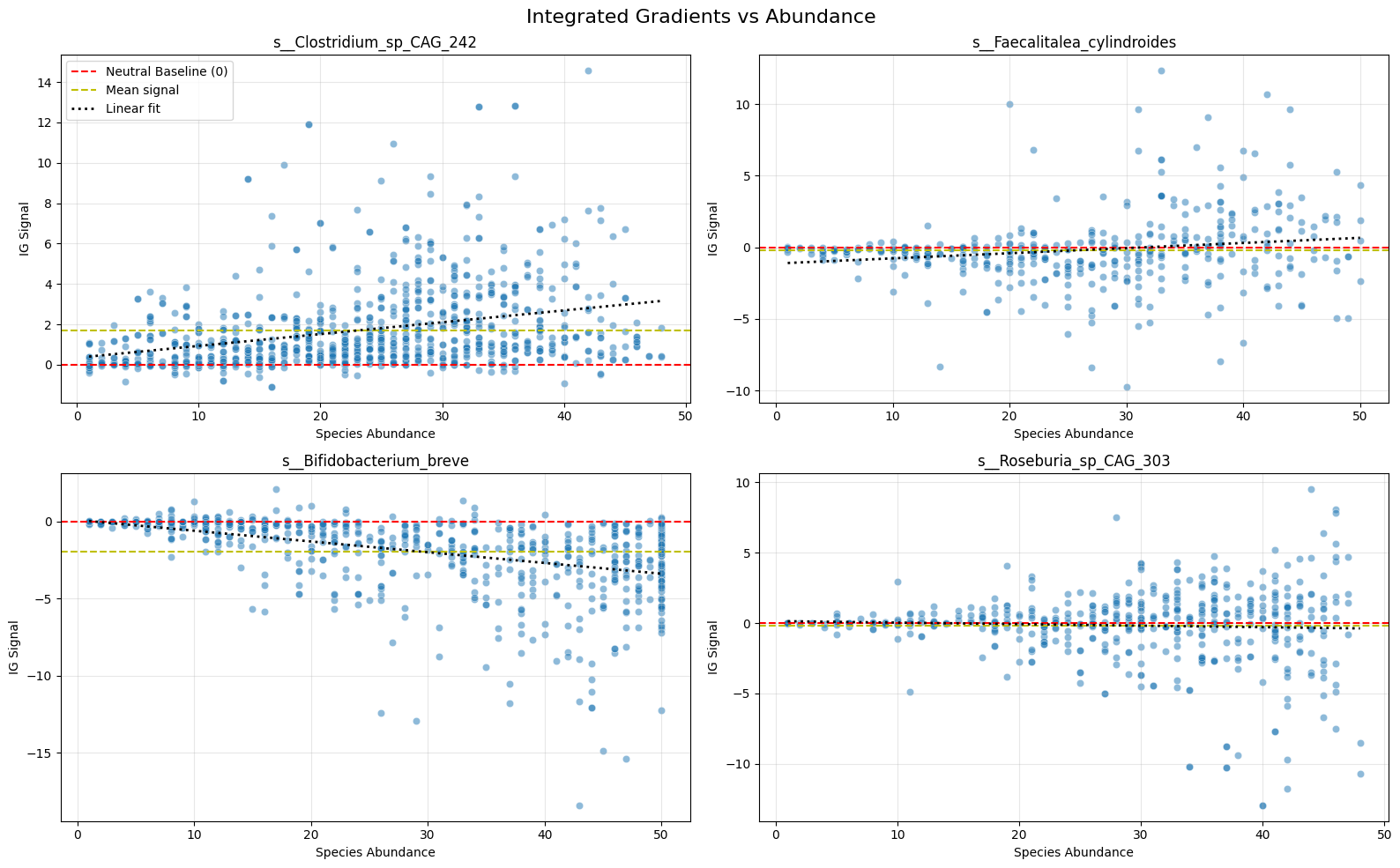}
\caption{
Abundance--attribution response profiles for four species. Each point represents one true positive or true negative sample. Larger attribution magnitudes at higher abundance levels are expected from the IG displacement factor \(A-A_0\): larger movement from baseline permits larger positive or negative attribution. Variation within the same abundance bin reflects sample level differences in path-averaged sensitivity, suggesting that the model's response to a species can depend on the surrounding microbiome context.
}
\label{fig:abundance_attribution_profiles}
\end{figure}

Figure~\ref{fig:abundance_attribution_profiles} plots \(IG_A\) attribution against observed abundance for individual species. The x-axis is the discrete binned abundance value, and the y-axis is the \(IG_A\) score for the IBD--Healthy logit contrast. Positive values increase the IBD margin, while negative values decrease it and support the healthy class.

The profiles show that abundance is not always treated as a linear or monotonic signal. Some species have consistently positive or negative attribution across abundance levels, while others show mixed behavior. This is information that a single unsigned attention score cannot provide: IG separates whether the model uses abundance as evidence for IBD or as evidence for healthy classification, and how that evidence changes across abundance regimes.

First-order IG identifies abundance regimes where a species contributes strongly to the model's IBD--Healthy margin, but it does not identify which other species may be modulating that response. The variation within abundance bins suggests that the attribution assigned to one species may depend on the broader microbiome context. This motivates a second-order extension for studying abundance--abundance interactions.

A natural second-order extension is grouped abundance--abundance interaction attribution exposing community rules. Let \(A_i \in \mathbb{R}^H\) denote the abundance embedding for species \(i\), let \(A'\) be the abundance baseline, and let
\[
\Delta A_i = A_i - A'_i.
\]
Following Integrated Hessians \cite{janizek2020explaining}, for \(i \neq j\), a token-level interaction score between species \(i\) and species \(j\) can be written as
\[
\Gamma^A_{i,j}
=
\Delta A_i^\top
\left[
\int_0^1
\int_0^1
\alpha\beta\,
\frac{
\partial^2 F\!\left(S, A' + \alpha\beta(A-A')\right)
}{
\partial A_i \partial A_j
}
\, d\alpha\, d\beta
\right]
\Delta A_j .
\]

Here, \(\partial^2 F / \partial A_i \partial A_j\) is the Hessian block between the abundance embeddings of species \(i\) and \(j\). This block measures how the model's sensitivity to the abundance of species \(i\) changes under infinitesimal changes to the abundance of species \(j\), and equivalently how species \(i\) modulates sensitivity to species \(j\) when the Hessian is symmetric. The left and right multiplication by \(\Delta A_i\) and \(\Delta A_j\) converts this coordinate-level Hessian block into a displacement-weighted token-level interaction score. Thus, \(\Gamma^A_{i,j}\) estimates the net interaction between moving species \(i\) and species \(j\) from their baseline abundances to their observed abundances along the IG path.

Because the Hessian can vary along the path, \(\Gamma^A_{i,j}\) should be interpreted as a path-averaged net interaction. A positive score indicates that the joint abundance movement tends to increase the IBD--Healthy logit contrast beyond the corresponding first-order effects, while a negative score indicates suppressive or health-supporting interaction. As in first-order interactions, if the interaction changes sign across abundance regimes, these effects may partially cancel in the integrated score; abundance-abundance interaction profiles are left as future work.

\section{Discussion and Limitations}

This study is not a clinical generalization benchmark. The reported validation accuracy establishes only that the model learned an IBD-discriminative signal suitable for attribution analysis; it should not be interpreted as evidence of clinical generalization. Evaluating broader microbiome generalization would require study-level holdout splits, as in Medearis et al. \cite{biomegpt2026}, as well as downstream validation through in silico and in vivo experiments.

Attribution should also not be interpreted as biological causality. The scores reported here are model-inferred evidence: they describe how the trained model uses species-abundance tokens to distinguish IBD from healthy samples. These patterns may suggest candidate disease-associated or health-associated species-abundance regimes, but causal claims require external validation through independent cohorts, perturbation experiments, and biological follow-up.

A limitation of standard IG is that it collapses the entire baseline to input path into one net attribution score. This is useful for completeness, but it can obscure conditional abundance-sensitivity structure. If the model's sensitivity to a species changes sign along the path from zero abundance to observed abundance, positive and negative regions may partially cancel in the final IG value. This does not invalidate the attribution; it means the score should be interpreted as net signed evidence over the chosen path.

The abundance--attribution response profiles partly address this issue by showing how attribution varies across observed abundance bins. Species with wide variation within the same abundance may indicate modulation by the surrounding sample-specific microbiome context. A natural extension is path-resolved attribution: instead of immediately integrating over all \(\alpha\), one can retain the sequence of path gradients
\[
g_A(\alpha)
=
\frac{\partial F(S,A_\alpha)}{\partial A}
\]
or compute segmented IG over subintervals \([u,v]\):
\[
IG_A^{[u,v]}
=
(A_v-A_u)\odot
\int_u^v
\frac{\partial F(S,A_\alpha)}{\partial A}
\,d\alpha.
\]
This would allow the model's signed abundance response to be studied across abundance levels, identifying where a species switches from IBD supporting to health supporting evidence, or where its effect is concentrated. Such path-resolved profiles may be useful for discovering conditional abundance rules, but they should be interpreted as diagnostics rather than replacements for full IG completeness over the baseline to input path.

This work studies only first-order attribution. First-order IG identifies species-abundance contributions to the IBD--Healthy margin, but it does not identify whether one species-abundance changes the model's sensitivity to another. Grouped Integrated Hessians \cite{janizek2020explaining} provide a natural future direction: instead of materializing an infeasible \((SH)\times(SH)\) Hessian over all species and embedding coordinates, coordinate-level interactions could be contracted into species-level abundance--abundance scores. This would allow analysis of whether the abundance of species \(j\) modulates the model's response to species \(i\). As in first-order IG, if the interaction changes sign across abundance regimes, these effects may partially cancel in the integrated score; abundance-conditioned interaction profiles are left as future work.

Finally, the current implementation computes IG in abundance-embedding space. A tokenizer-aware path through the learned abundance encoder may be worth exploring in future work, but the present study focuses on comparing signed IG attribution against unsigned \texttt{[CLS]} attention.

\section{Conclusion}
This work shows that \texttt{[CLS]} attention weights are insufficient for explaining BiomeGPT-style microbiome decisions. It ranks fused species--abundance tokens, but cannot distinguish polarity \cite{liu2022rethinking} between evidence supporting IBD from evidence supporting health. It cannot isolate the effects of source inputs that compose a token on the model output prediction.

Integrated Gradients addresses this limitation by attributing the logit contrast
\[
z_{\mathrm{IBD}}-z_{\mathrm{Healthy}}.
\]
Positive attribution supports IBD, while negative attribution supports healthy classification. The source-derived baseline
\[
T'=S+A_0
\]
preserves species identity and attributes the change from zero to observed abundance, giving fused-token IG a direct biological interpretation.

Empirically, IG reveals directional evidence and abundance--attribution response patterns unavailable from attention rankings alone. Because BiomeGPT is pretrained on large microbiome datasets and then adapted to disease-specific tasks, these attributions may also help probe how reusable microbial structure learned during pretraining is expressed and specialized during fine-tuning.

Overall, \texttt{[CLS]} attention weights remain useful for describing internal representation structure, but fusion-aware IG provides a stronger basis for understanding how observed microbial abundances influence the model's disease--health decision.

{
\small
\bibliographystyle{plainnat}
\bibliography{references}

\begin{thebibliography}{17}
\providecommand{\natexlab}[1]{#1}
\providecommand{\url}[1]{\texttt{#1}}
\expandafter\ifx\csname urlstyle\endcsname\relax
  \providecommand{\doi}[1]{doi: #1}\else
  \providecommand{\doi}{doi: \begingroup \urlstyle{rm}\Url}\fi

\bibitem[Ancona et~al.(2018)Ancona, Ceolini, {\"O}ztireli, and
  Gross]{ancona2018towards}
Marco Ancona, Enea Ceolini, Cengiz {\"O}ztireli, and Markus Gross.
\newblock Towards better understanding of gradient-based attribution methods
  for deep neural networks.
\newblock In \emph{International Conference on Learning Representations}, 2018.
\newblock URL \url{https://arxiv.org/abs/1711.06104}.
\newblock arXiv:1711.06104.

\bibitem[Blanco-M{\'i}guez et~al.(2023)Blanco-M{\'i}guez, Beghini, Cumbo,
  McIver, Thompson, Zolfo, Manghi, Dubois, Huang, Thomas, Nickols, Piccinno,
  Piperni, Pun{\v{c}}och{\'a}{\v{r}}, Valles-Colomer, Tett, Giordano, Davies,
  Wolf, Berry, Spector, Franzosa, Pasolli, Asnicar, Huttenhower, and
  Segata]{blanco2023metaphlan4}
Aitor Blanco-M{\'i}guez, Francesco Beghini, Fabio Cumbo, Lauren~J. McIver,
  Kelsey~N. Thompson, Moreno Zolfo, Paolo Manghi, Leonard Dubois, Kun~D. Huang,
  Andrew~Maltez Thomas, William~A. Nickols, Gianmarco Piccinno, Elisa Piperni,
  Michal Pun{\v{c}}och{\'a}{\v{r}}, Mireia Valles-Colomer, Adrian Tett,
  Francesca Giordano, Richard Davies, Jonathan Wolf, Sarah~E. Berry, Tim~D.
  Spector, Eric~A. Franzosa, Edoardo Pasolli, Francesco Asnicar, Curtis
  Huttenhower, and Nicola Segata.
\newblock Extending and improving metagenomic taxonomic profiling with
  uncharacterized species using {MetaPhlAn} 4.
\newblock \emph{Nature Biotechnology}, 41:\penalty0 1633--1644, 2023.
\newblock \doi{10.1038/s41587-023-01688-w}.
\newblock URL \url{https://doi.org/10.1038/s41587-023-01688-w}.

\bibitem[Cui et~al.(2024)Cui, Wang, Maan, Pang, Luo, Duan, and
  Wang]{cui2024scgpt}
Haotian Cui, Chloe Wang, Hassaan Maan, Kuan Pang, Fengning Luo, Nan Duan, and
  Bo~Wang.
\newblock {scGPT}: toward building a foundation model for single-cell
  multi-omics using generative {AI}.
\newblock \emph{Nature Methods}, 21:\penalty0 1470--1480, 2024.
\newblock \doi{10.1038/s41592-024-02201-0}.
\newblock URL \url{https://doi.org/10.1038/s41592-024-02201-0}.

\bibitem[Devlin et~al.(2019)Devlin, Chang, Lee, and Toutanova]{devlin2019bert}
Jacob Devlin, Ming-Wei Chang, Kenton Lee, and Kristina Toutanova.
\newblock Bert: Pre-training of deep bidirectional transformers for language
  understanding.
\newblock In \emph{Proceedings of the 2019 conference of the North American
  chapter of the association for computational linguistics: human language
  technologies, volume 1 (long and short papers)}, pages 4171--4186, 2019.

\bibitem[Gorishniy et~al.(2021)Gorishniy, Rubachev, Khrulkov, and
  Babenko]{ft_transformer_2021}
Yury Gorishniy, Ivan Rubachev, Valentin Khrulkov, and Artem Babenko.
\newblock Revisiting deep learning models for tabular data.
\newblock In \emph{Advances in Neural Information Processing Systems
  (NeurIPS)}, volume~34, pages 18932--18943, 2021.
\newblock URL
  \url{https://proceedings.neurips.cc/paper/2021/hash/9d86d83f925f2149e9edb0ac3b49229c-Abstract.html}.

\bibitem[Jain and Wallace(2019)]{jain2019attention}
Sarthak Jain and Byron~C. Wallace.
\newblock Attention is not explanation.
\newblock In \emph{Proceedings of the 2019 Conference of the North American
  Chapter of the Association for Computational Linguistics}, pages 3543--3556,
  2019.
\newblock URL \url{https://arxiv.org/abs/1902.10186}.
\newblock arXiv:1902.10186.

\bibitem[Janizek et~al.(2020)Janizek, Sturmfels, and
  Lee]{janizek2020explaining}
Joseph~D. Janizek, Pascal Sturmfels, and Su-In Lee.
\newblock Explaining explanations: Axiomatic feature interactions for deep
  networks.
\newblock \emph{arXiv preprint arXiv:2002.04138}, 2020.
\newblock \doi{10.48550/arXiv.2002.04138}.
\newblock URL \url{https://arxiv.org/abs/2002.04138}.
\newblock arXiv:2002.04138.

\bibitem[Kokhlikyan et~al.(2020)Kokhlikyan, Miglani, Martin, Wang, Alsallakh,
  Reynolds, Melnikov, Kliushkina, Araya, Yan, and Reblitz-Richardson]{captum}
Narine Kokhlikyan, Vivek Miglani, Miguel Martin, Edward Wang, Bilal Alsallakh,
  Jonathan Reynolds, Alexander Melnikov, Natalia Kliushkina, Carlos Araya, Siqi
  Yan, and Orion Reblitz-Richardson.
\newblock Captum: A unified and generic model interpretability library for
  pytorch, 2020.
\newblock URL \url{https://arxiv.org/abs/2009.07896}.
\newblock arXiv:2009.07896.

\bibitem[Liu et~al.(2022)Liu, Li, Guo, Kong, Li, and Wang]{liu2022rethinking}
Yibing Liu, Haoliang Li, Yangyang Guo, Chenqi Kong, Jing Li, and Shiqi Wang.
\newblock Rethinking attention-model explainability through faithfulness
  violation test.
\newblock In \emph{Proceedings of the 39th International Conference on Machine
  Learning}, pages 13858--13871. PMLR, 2022.

\bibitem[Medearis et~al.(2026)Medearis, Zhu, and Zomorrodi]{biomegpt2026}
Nicholas~A. Medearis, Siyao Zhu, and Ali~R. Zomorrodi.
\newblock Biomegpt: A foundation model for the human gut microbiome.
\newblock \emph{bioRxiv}, 2026.
\newblock \doi{10.64898/2026.01.05.697599}.
\newblock URL \url{https://doi.org/10.64898/2026.01.05.697599}.

\bibitem[Pasolli et~al.(2017)Pasolli, Schiffer, Manghi, Renson, Obenchain,
  Truong, Beghini, Malik, Ramos, Dowd, Huttenhower, Morgan, Segata, and
  Waldron]{pasolli2017curatedmetagenomicdata}
Edoardo Pasolli, Lucas Schiffer, Paolo Manghi, Audrey Renson, Valerie
  Obenchain, Duy~Tin Truong, Francesco Beghini, Faizan Malik, Marcel Ramos,
  Jennifer~B. Dowd, Curtis Huttenhower, Martin Morgan, Nicola Segata, and Levi
  Waldron.
\newblock Accessible, curated metagenomic data through {ExperimentHub}.
\newblock \emph{Nature Methods}, 14:\penalty0 1023--1024, 2017.
\newblock \doi{10.1038/nmeth.4468}.
\newblock URL \url{https://doi.org/10.1038/nmeth.4468}.

\bibitem[Pope et~al.(2025)Pope, Varma, Tataru, David, and
  Fern]{pope2025microbiome_language_model}
Quintin Pope, Rohan Varma, Christine Tataru, Maude~M. David, and Xiaoli Fern.
\newblock Learning a deep language model for microbiomes: The power of large
  scale unlabeled microbiome data.
\newblock \emph{PLOS Computational Biology}, 21\penalty0 (5):\penalty0
  e1011353, 2025.
\newblock \doi{10.1371/journal.pcbi.1011353}.
\newblock URL \url{https://doi.org/10.1371/journal.pcbi.1011353}.

\bibitem[Roy and Kundu(2024)]{roy2024udig}
Swarnava~Sinha Roy and Ayan Kundu.
\newblock Uniform discretized integrated gradients: An effective attribution
  based method for explaining large language models, 2024.
\newblock URL \url{https://arxiv.org/abs/2412.03886}.
\newblock arXiv:2412.03886.

\bibitem[Sundararajan et~al.(2017)Sundararajan, Taly, and
  Yan]{sundararajan2017axiomatic}
Mukund Sundararajan, Ankur Taly, and Qiqi Yan.
\newblock Axiomatic attribution for deep networks.
\newblock In \emph{International Conference on Machine Learning}, pages
  3319--3328, 2017.
\newblock URL \url{https://arxiv.org/abs/1703.01365}.
\newblock arXiv:1703.01365.

\bibitem[Theodoris et~al.(2023)Theodoris, Xiao, Chopra, Chaffin, Al~Sayed,
  Hill, Mantineo, Brydon, Zeng, Liu, and Ellinor]{theodoris2023geneformer}
Christina~V. Theodoris, Ling Xiao, Anant Chopra, Mark~D. Chaffin, Zeina~R.
  Al~Sayed, Matthew~C. Hill, Helene Mantineo, Elizabeth~M. Brydon, Zexian Zeng,
  X.~Shirley Liu, and Patrick~T. Ellinor.
\newblock Transfer learning enables predictions in network biology.
\newblock \emph{Nature}, 618\penalty0 (7965):\penalty0 616--624, 2023.
\newblock \doi{10.1038/s41586-023-06139-9}.
\newblock URL \url{https://doi.org/10.1038/s41586-023-06139-9}.

\bibitem[Wiegreffe and Pinter(2019)]{wiegreffe2019attention}
Sarah Wiegreffe and Yuval Pinter.
\newblock Attention is not not explanation.
\newblock In \emph{Proceedings of the 2019 Conference on Empirical Methods in
  Natural Language Processing}, pages 11--20, 2019.
\newblock URL \url{https://arxiv.org/abs/1908.04626}.
\newblock arXiv:1908.04626.

\bibitem[Yang et~al.(2022)Yang, Wang, Wang, Fang, Tang, Huang, Lu, and
  Yao]{yang2022scbert}
Fan Yang, Wenchuan Wang, Fang Wang, Yuan Fang, Duyu Tang, Junzhou Huang, Hui
  Lu, and Jianhua Yao.
\newblock {scBERT} as a large-scale pretrained deep language model for cell
  type annotation of single-cell {RNA}-seq data.
\newblock \emph{Nature Machine Intelligence}, 4:\penalty0 852--866, 2022.
\newblock \doi{10.1038/s42256-022-00534-z}.
\newblock URL \url{https://doi.org/10.1038/s42256-022-00534-z}.

\end{thebibliography}
}
\end{document}